\documentclass[10pt,twocolumn,letterpaper]{article}

\usepackage{wacv}              

\usepackage{graphicx}
\usepackage{amsmath}
\usepackage{amssymb}
\usepackage{booktabs}
\usepackage{wacv}
\usepackage{times}
\usepackage{epsfig}
\usepackage{algorithm2e}
\usepackage{floatrow}
\RestyleAlgo{ruled}
\usepackage{tabularx}
\usepackage{multirow}
\usepackage{svg}
\usepackage{lipsum}
\usepackage{caption}
\usepackage{wrapfig}
\usepackage{amssymb}
\usepackage{pifont}
\newcommand{\xmark}{\ding{55}}%
\makeatletter
\let\@fnsymbol\@arabic
\makeatother

\usepackage[pagebackref,breaklinks,colorlinks]{hyperref}

\usepackage[capitalize]{cleveref}
\crefname{section}{Sec.}{Secs.}
\Crefname{section}{Section}{Sections}
\Crefname{table}{Table}{Tables}
\crefname{table}{Tab.}{Tabs.}

\def\wacvPaperID{8} 
\def\confName{WACV}
\def\confYear{2025}

\begin{document}

\title{Segment Anything in Light Fields for Real-Time Applications via Constrained Prompting}

\author{
    Nikolai Goncharov\thanks{\href{mailto:ngon2624@uni.sydney.edu.au}{ngon2624@uni.sydney.edu.au}} \qquad Donald G. Dansereau\thanks{\href{mailto:donald.dansereau@sydney.edu.au}{donald.dansereau@sydney.edu.au}}\\\\University of Sydney\thanks{Australian Centre For Robotics (ACFR), School of Aerospace, Mechanical and Mechatronic Engineering, The University of Sydney, 2006 NSW, Australia.}
}
\maketitle

\begin{abstract}
   Segmented light field images can serve as a powerful representation in many of computer vision tasks exploiting geometry and appearance of objects, such as object pose tracking. In the light field domain, segmentation presents an additional objective of recognizing the same segment through all the views. Segment Anything Model 2 (SAM 2) allows producing semantically meaningful segments for monocular images and videos. However, using SAM 2 directly on light fields is highly ineffective due to unexploited constraints. In this work, we present a novel light field segmentation method that adapts SAM 2 to the light field domain without retraining or modifying the model. By utilizing the light field domain constraints, the method produces high quality and view-consistent light field masks, outperforming the SAM 2 video tracking baseline and working 7 times faster, with a real-time speed. We achieve this by exploiting the epipolar geometry cues to propagate the masks between the views, probing the SAM 2 latent space to estimate their occlusion, and further prompting SAM 2 for their refinement. Code and additional materials are available at \small{\href{https://roboticimaging.org/Projects/LFSAM/}{https://roboticimaging.org/Projects/LFSAM/}}.
\end{abstract}

\section{Introduction}
One of the extensions of the regular monocular images are the light field images. They are generalized images that capture the 4D subset of the plenoptic function and allow understanding of the scene on the level of rays \cite{levoy2023light}, \cite{wu2017light}. They are already proven to provide strong cues for depth estimation \cite{wang2015occlusion}, \cite{mishiba2020fast}, structure from motion \cite{johannsen2015linear}, \cite{tsai2018distinguishing} and odometry \cite{dansereau2011plenoptic}, \cite{zeller2017calibration}. Light fields are also tightly related to recent advancements in novel view rendering \cite{mildenhall2021nerf}, \cite{gao2022nerf}. 

Image segmentation connects sensor observations with objects on the scene. It is applied in robotics, autonomous driving, scene understanding, medicine and many more domains. Segmenting objects on a light field image provides powerful representations: for example, for dynamic or cluttered scene motion understanding. Segmented light fields can be used as a replacement of CAD models, NeRF and other heavy representations used in tasks such as zero-shot model free 6D object pose tracking \cite{wen2024foundationpose} as means of supervision. Those methods prove to be promising in such important domains as autonomous driving \cite{di2024zero123}. Existing light field segmentation methods \cite{khan2019view} \cite{hamad2021alfo} already provide high performance and view-consistent results, however they focus on fixed sized superpixel segmentation, independent of the scene semantics and understanding of object/instance boundaries.

Recently, there's been an increased effort in creating large foundation models trained on industrial scale and applied in natural language processing and computer vision \cite{awais2023foundational}. As an example, Segment Anything Model (SAM) is a foundation model that produces high quality promptable segmentation \cite{kirillov2023segment} for monocular images. Adapting such a model to the light field domain through fine-tuning is challenging due to the low amount of existing light field data and its high dimensionality. Segment Anything Model 2 (SAM 2) \cite{ravi2024sam} introduced an improved version of the image segmentation model, as well as a model able to propagate segments across videos. This can be directly applied to segment a light field with impressive quality. However, treating a light field as a regular video does not leverage the rich cues and constraints of its 4D structure.

In this work, we extend the capabilities of the SAM 2 promptable segmentation to work on light field images. No retraining, fine-tuning, or modifying the network is required. We combine the SAM 2 powerful semantic understanding with epipolar geometry constraints. As a result, we get a subview-consistent light field segmentation, while preserving the visual quality of the original model. We initialize our method with automatic mask generation and compare against the baseline of SAM 2 video segment tracking model. To summarize, we make the following contributions:
\begin{itemize}
    \setlength\itemsep{0.1em}
    \item \textbf{A novel method for light field image segmentation}, extending the capabilities of SAM 2 to a novel modality without modification or retraining.
    \item \textbf{Segmentation refinement}, a two-step method of light field segmentation with an option to fall back on a simpler but effective segmentation option for uncertain cases. The first step exploits rich light field cues to obtain coarse light field masks, and the second step utilizes the SAM 2 prompted segmentation to apply light field constraints for the result refinement.
    \item \textbf{Semantic occluding}, a technique that uses latent semantic features of SAM's MAE \cite{he2022masked} image encoder model to estimate the occluded regions of light field segments to refine the prompts provided to the model.
\end{itemize}

We show that our method produces semantically accurate and spatio-angularly consistent segments, avoids excessive oversegmentation of objects, achieves higher performance than SAM 2 video tracking, while being $7$ times faster. This work establishes a way to generalize a pretrained foundation model to a new modality, allowing effective segmentation of large baseline light field images and opening a door for generalizing other 2D modality methods to be extended to light field images. We will release the code for the method on acceptance of the paper.

\textbf{Limitations.} We are using the middle subview as a reference to initiate the segmentation, therefore the objects occluded in this subview are ignored. This is also true for the video tracking baseline. We use gradient-based depth estimation to propagate the segments through the subviews, which may cause decreased performance dealing with reflective and refractive surfaces. We inherit all the strengths, but also the weaknesses of the domain SAM 2 image model was trained on. For example, low light conditions.

\section{Related work}
\textbf{Foundation models.} Segment Anything (SAM) \cite{kirillov2023segment} is a foundation model for prompt-based zero shot segmentation on a large variety of domains. It received an extension to a lot of domains such as video segment tracking \cite{yang2023track}, medical images \cite{ma2024segment} and open-set tasks \cite{zou2024segment}. There's still a lot of potential to extend SAM to novel modalities. Segment Anything Model 2 (SAM 2) \cite{ravi2024sam} extended SAM to the video domain, as well as improving the image segmentation model.

\textbf{Light fields.} Light fields have been applied to variety of tasks such as depth estimation \cite{wang2015occlusion}, \cite{mishiba2020fast}, structure from motion \cite{johannsen2015linear}, \cite{tsai2018distinguishing}. Moreover, they have been proved useful in estimating odometry \cite{dansereau2011plenoptic}, \cite{zeller2017calibration} and scene flow   \cite{srinivasan2015oriented}, \cite{ma20183d}, which are powerful representations in robotics and autonomous driving. However, those methods are dense and do not benefit from the prior constraints that can be obtained through representations such as segmentation using modern models. There's work presenting positive examples of implementing such constraints in monocular and stereo scene flow estimation \cite{ma2019deep}, \cite{li2021neural}, which can later be extended to light fields.

\textbf{Light field segmentation.} The task of segmentation has also been extended to light fields, superpixel segmentation. Hog \etal \cite{hog2017superrays} extend monocular superpixels to light field superrays and their estimation.  Zhu \etal \cite{zhu20174d} introduce a self-similarity measure for light field superpixels, showing their invariance to refocus. VCLS \cite{khan2019view} is a method that improves the estimation of the light field superpixels by propagating the segmentation to more views, making it more robust to occlusion and consistent across the views. Lv \etal \cite{lv20204d} use the light field superpixels combined with disparity and user scribbles to obtain a more accurate segmentation of a light field. The main limitation of the light field segmentation methods is either strong oversegmentation or the requirement in user input or other additional representations such as depth or disparity. Additionally, the recent advancements in foundation models have not yet been leveraged for the task. Our method utilizes the foundation models and produces semantically meaningful segments without any additional representations needed.

\textbf{6D object pose tracking.} The methods that track the object's pose on a cluttered scene rely either on knowing the object's 3D models \cite{xiang2017posecnn}, or pre-recording an explicit representation such as NeRF \cite{wen2024foundationpose} for supervising the estimation. Those methods are showing to be promising for such important applications as autonomous driving \cite{di2024zero123} and, therefore, are in need of real-time supervision. Light fields provide rich cues about the object's 6D motion if it's the only motion present on the scene. Therefore, those heavy representations can be replaced with light field segments of those objects.

\begin{figure*}[t]
    \centering
    \includegraphics[width=\linewidth,height=15\baselineskip]{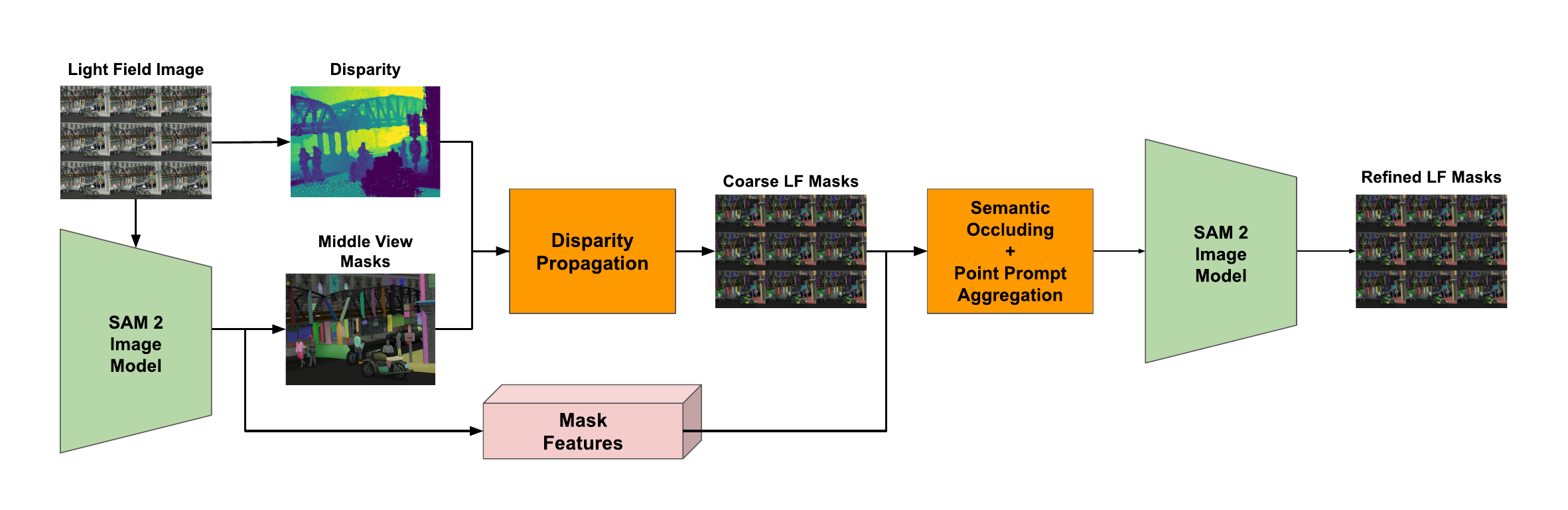}
    \captionof{figure}{Our method. First, we obtain middle view segmentation and a disparity map for a light field image. We perform disparity propagation on the middle view masks, resulting in coarse mask predictions. We occlude those predictions using mask feature similarities with respect to the source segment. Then, we aggregate the resulting masks into points and use them prompt SAM 2 image model in the rest of the subviews, resulting in the refined light field mask predictions.}
    \label{fig:pipeline}
\end{figure*}

\section{Method}
The structure of our method is presented in Figure \nolinebreak \ref{fig:pipeline}. Given a light field image, we leverage the SAM 2 image model to obtain \textit{source} object masks from the middle subview. We exploit the epipolar geometry constraints on the structure of a light field image and use disparity to propagate the mask to the rest of the views, obtaining \textit{coarse} mask position predictions. Additionally, we utilize the SAM 2 image encoder to obtain a per-pixel semantic latent feature vector for both the source mask and all coarse predicted masks. We use cosine similarities to remove occluded pixels from the coarse predicted masks, which can then be used as prompts for SAM2. However, we have found precise single point prompts to be the most effective in practice. Therefore, for each coarse prediction mask, we prompt SAM 2 with its centroid point and bounding box to obtain a refined prediction in the corresponding subview, eventually obtaining the result light field mask.

To summarize, we obtain a light field segment using the following steps: initial segmentation, disparity propagation, semantic occluding and segmentation refinement.

\subsection{Initial segmentation}
Let $L[s, t, u, v, c] \in \mathbb{R}^{S \times T \times U \times V \times 3}$ be a light field image, with subview dimensions $[s, t]$ and spatial dimensions $[u, v]$. Given a prompt $P[s_m, t_m]$ in the middle subview, which can be a point or a box, our goal is to obtain the corresponding light field mask $M[s, t, u, v]$. As a first step, we pick the middle subview $L[s_m, t_m]$ and decode the source mask:
\begin{equation} \label{init-seg}
    M[s_m, t_m] = \mathrm{SAM2_{img}}(L[s_m, t_m], P[s_m, t_m]).
\end{equation}

\subsection{Disparity propagation}
We process $L$ to get a disparity map $d \in \mathbb{R}^{U \times V}$ using a method based on the structure tensor \cite{wanner2013variational}. The goal of disparity propagation is obtaining $M_{\mathrm{coarse}}[s, t, u, v]$, a coarse version of the predicted light field mask, which doesn't take occlusion into account. Let $(s_m, t_m, u_m, v_m) \in \mathbb{R}^{N \times 4}$ be the set of points on the source mask and $(s_i, t_i, u_i, v_i) \in \mathbb{R}^{N \times 4}$ -- on the predicted mask in the subview $[i, j]$. To predict the position of $M[s_m, t_m]$ in subview $[i, j]$, we linearly project the points:
\begin{equation}
    \begin{bmatrix} u_i \\ v_i \end{bmatrix} = \begin{bmatrix} u_m \\ v_m \end{bmatrix} + d[u_m, v_m] \cdot \begin{bmatrix} s_m - i \\ t_m - j \end{bmatrix}.
\end{equation}
We repeat the procedure for every subview to find $M_{\mathrm{coarse}}[s, t, u, v]$.

\subsection{Semantic occluding}
Parts of the source mask might be occluded by the objects at the front of the scene in some of the subviews of the target light field mask. This isn't taken into account by the coarse prediction, since we directly project it. We can remove the occluded pixels from $M_{\mathrm{coarse}}[s, t, u, v]$ using semantic features. The assumption we make is that the occluder has different semantic properties.

Given a source $M[s_m, t_m]$ and its subview image $L[s_m, t_m]$, we estimate its latent feature vector $\mathcal{F}(M[s_m, t_m]) \in \mathbb{R}^K$. SAM 2 uses MAE \cite{he2022masked} to split images into $P \times P$ patches and encode by a feature extractor to get a feature map $\mathcal{F} \in \mathbb{R}^{P \times P \times K}$, where $K$ is the size of MAE's embedding. To get $\mathcal{F}(M[s_m, t_m])$, we upscale the feature map of $L[s_m, t_m]$, perform pointwise multiplication with $M[s_m, t_m]$, and average all non-zero $K$-dimensional entries.

We obtain $\mathcal{F}(M_{\mathrm{coarse}}[i, j]) \in \mathbb{R}^{N \times K}$ similarly, but instead of averaging we leave it in a per-pixel form, where $N$ is the number of pixels in the mask. We then compute cosine similarities of $\mathcal{F}(M_{\mathrm{coarse}}[i, j])$ and $\mathcal{F}(M[s_m, t_m])$ and drop the points from $M[s_m, t_m]$ that are below a certain threshold $t_{\mathrm{sim}}$.

\subsection{Segmentation refinement}
To further refine the predictions, we reprompt the model in all the subviews with the coarse masks $M_{\mathrm{coarse}}[s, t]$. However, the coarse mask might still include regions out of the boundaries of the initial object, due to the remaining occlusion and improper disparity values. Using it as a prompt might unite it with undesirable objects. Instead, in each subview, we prompt the model with a single precise prompt: the centroid of the mask $\mathcal{C}(M_{\mathrm{coarse}}[s, t]) \in \mathbb{R}^2$. Additionally, we find a bounding box around the mask $\mathcal{B}(M_{\mathrm{coarse}}[s, t]) \in \mathbb{R}^4$. Our prompt is then the union of the point and the box:
\begin{equation}
    P[s, t] = \{\mathcal{B}(M_{\mathrm{coarse}}[s, t]), \mathcal{C}(M_{\mathrm{coarse}}[s, t])\}.
\end{equation}
As a result, we obtain the refined light field mask prediction $M[s, t, u, v]$:
\begin{equation} \label{init-seg}
    M[s, t] = \mathrm{SAM2_{img}}(L[s, t], P[s, t])).
\end{equation}

Finally, we estimate Intersection Over Union (IoU) between $M[s, t]$ and $M_{\mathrm{coarse}}[s, t]$ and we fall back on $M_{\mathrm{coarse}}[s, t]$ if it's below a certain threshold $t_{\mathrm{IoU}}$ in case SAM 2 doesn't produce refined segments containing mutual boundaries with the coarse ones, or the prompt falls onto an incorrect object, making our segmentation naturally adapt to challenging situations.

\section{Experiments}
We use 40 synthetic light fields rendered in Blender, sized $9 \times 9$ from the validation part of the UrbanLF-Synthetic \cite{sheng2022urbanlf} dataset. Since there's no way to uniquely define a segment provided by prompted segmentation, we use a set of metrics for light field superpixel segmentation, allowing us to utilize the ground truth semantic labels provided by the dataset. Both labels and disparity maps in this dataset are provided for all the views, allowing us to compute both segmentation accuracy and cross-view consistency metrics.

\textbf{Implementation details.}
For both our method and the baseline, we use a Hiera small version of SAM 2 \cite{ravi2024sam} in the automatic mask generation mode with $64$ points per side. For the baseline method, we use the video tracking mode of SAM 2, segmenting the first view, reshaping the light field image in a snake-like pattern, and tracking the segments. For our method, we use $t_{\mathrm{sim}} = 0.7$, dropping high-certainty occluded points, while leaving enough points to sample centroids, which we additionally weight by the similarities. We find $t_{\mathrm{IoU}} = 0.1$ enough to ensure common boundaries between coarse and fine masks. The method is implemented in PyTorch and benchmarked on a single Tesla V100 with 30 GB total memory.

\subsection{Metrics} \label{sec:metrics}
\textbf{Cross-view consistency metrics.} 
For each light field mask, we first backproject the masks from each view into the reference view using the ground truth disparity for efficient pixelwise comparison. From this representation, we compute the following set of metrics:
\begin{itemize}
    \item \textbf{Self IoU (SIoU)} — similar to Self Similarity introduced in \cite{zhu20174d}, but instead of mean centroid distance, mean IoU is computed between the projected segments.
    \item \textbf{Labels Per Pixel (LPP)} — introduced in \cite{khan2019view} as additional consistency metric. Computes the number of unique segments assigned to a single pixel.
\end{itemize}

\textbf{Segmentation metrics}
\begin{itemize}
    \item \textbf{Achievable Accuracy (AA)} — introduced in \cite{liu2011entropy} for superpixel segmentation. Assigns each superpixel to a ground truth segment with the highest amount of mutual pixels, and calculates accuracy between the two output segmentation masks.
    \item \textbf{Undersegmentation Error (UE)} — keeps track of the predicted segments overflowing through the boundaries of the ground truth segments. We use the version described in \cite{neubert2012superpixel}.
\end{itemize}

\textbf{Performance metrics}
\begin{itemize}
    \item \textbf{Coverage (C)} — the ratio of unsegmented pixels on the light field. This metric is useful to compare against baseline, to understand where our method loses segments.
    \item \textbf{Computational time (T)} — provided in milliseconds per mask per subview.
\end{itemize}

Those metrics provide a clear performance picture for a downstream task such as 6D object pose tracking in autonomous driving. The segment shouldn't contain any extraneous objects or their parts, which is the most desirable property for this task, since the motion between different object instances on a cluttered scene can differ. \textbf{SIoU}, \textbf{AA} and \textbf{UE} are the most relevant metrics for this task, while \textbf{computational time} is relevant for the real-time aspect.

\subsection{Quantitative results}
Quality metrics comparison for UrbanLF-Synthetic dataset are shown in Table \ref{table:1}. We report the average value of each metric across all 40 scenes in the dataset. We compare SAM 2 video (baseline) against our method. We find that our method runs $7$ times faster, while outperforming or the baseline on all the consistency and segmentation metrics of the benchmark. As a result, obtaining a mask in all views from a single prompt takes less than 1 second, which is feasible for real-time segmentation. Both the speedup and the quality improvement are achieved by directly providing the model with geometry priors through prompts.

\subsection{Qualitative results}
\begin{figure*}
    \centering
    \includegraphics[width=500pt]{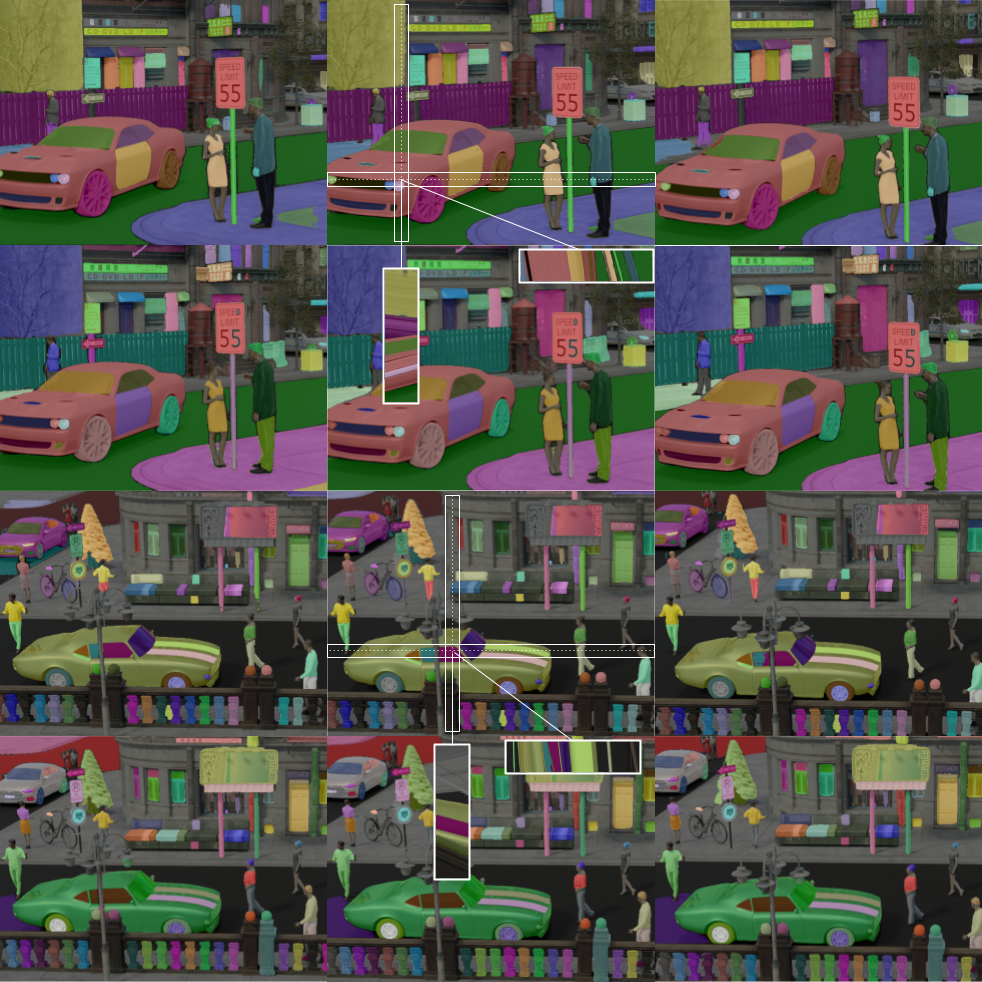}
    \caption{Qualitative results for our method on two scenes from UrbanLF \cite{sheng2022urbanlf} dataset. Our method is at the top row, and SAM 2 video tracking is at the bottom. Top left, middle and bottom subviews are visualized left-to-right. In each area highlighted by a rectangle, an epipolar plane image is taken across the yellow dotted line, upsampled along the subview axis and visualized.}
    \label{fig:qual2}
\end{figure*}

The comparative qualitative results are presented in Figure \ref{fig:qual2}. We visualize the border subviews for two light field scenes from UrbanLF-Synthetic dataset. Our method doesn't deform the shapes of segments, and properly matches them from the middle to the border subviews. The epipolar images show that even for most objects, our labeling is uninterrupted in all views. For the bottom image, an inconsistency can be noticed in the labels of two car windows, challenging objects due to their reflective nature. Please refer to supplementary videos for clearer visualization.

\subsection{Ablation studies}
Ablation studies for UrbanLF-Synthetic dataset are shown in Table \ref{table:2}. We study the effect of segmenation refinement ($\mathrm{ref}$ in the table) and semantic occluding ($\mathrm{occ}$ in the table). The most basic option doesn't handle occlusions and objects for which the depth estimation method fails. Adding $\mathrm{ref}$ and $\mathrm{occ}$ is shown to improve the metrics. The difference is clear in the qualitative ablations, provided in the supplementary materials video.

\section{Conclusion}
We introduced a novel light field segmentation method for fast and efficient large scale light field segmentation. Additionally, we introduced a segmentation refinement technique that exploits rich epipolar constraints. Finally, we introduced an occlusion estimating technique based on semantic features. We've leveraged the potential of a foundation model in a new modality  without retraining. The method produces semantically meaningful segments while preserving cross-subview consistency. We've shown the performance of our method on large-baseline data excels that of SAM 2 video tracking model, while being $7$ times faster.

In future work, there is scope to fully adapt the task of promptable segmentation to light fields: having an option to prompt the model with rays from any view to find the light field mask. Additionally, adapting the method to reflective and refractive surfaces by modifying the exploited disparity estimation. Moreover, overcoming the SAM 2 limitations remains a challenge: for example, low light conditions and limited domains. We also find that the visual quality of the baseline is very impressive. We recognize the challenge of finding additional benchmarks to see where our method can be improved. The output segmentation method is to be applied to downstream tasks such as 6D object pose tracking for autonomous driving.

\section{Acknowledgements}
This research was supported in part by funding from Ford Motor Company.

\begin{table}
\centering
\footnotesize
\begin{tabular}{c c c c c c c} 
 \hline
 Method & SIoU$\uparrow$ & LPP$\downarrow$ & AA$\uparrow$ & UE$\downarrow$ & C$\uparrow$ & T$\downarrow$ \\ [0.5ex] 
 \hline
 \hline
 \\[-1em]
 baseline & 0.765 & 1.452 & 0.970 & 0.038 & \textbf{0.329} & 108.5 \\
 \hline
 \\[-1em]
 ours & \textbf{0.768} & \textbf{1.408} & \textbf{0.973} & \textbf{0.032} & 0.309 & \textbf{15.2} \\
 \hline
\end{tabular}
\caption{UrbanLF Synthetic \cite{sheng2022urbanlf} dataset quantitative metrics comparison. Best result is shown with bold. Our method outputs comparable quality while significantly increasing computational speed.}
\label{table:1}
\end{table}

\begin{table}
\centering
\footnotesize
\begin{tabular}{c c c c c c c c} 
 \hline
 $\mathrm{ref}$ & $\mathrm{occ}$ & SIoU$\uparrow$ & LPP$\downarrow$ & AA$\uparrow$ & UE$\downarrow$ & C$\uparrow$ & T$\downarrow$ \\ [0.5ex] 
 \hline
 \hline
 \\[-1em]
 \xmark & \xmark & \textcolor{red}{\textbf{0.734}} & 1.414 & \textcolor{red}{\textbf{0.949}} & \textcolor{red}{\textbf{0.052}} & 0.303 & \textbf{3} \\
 \hline
 \\[-1em]
 $\checkmark$ & \xmark & 0.764 & 1.414 & 0.965 &  0.040 & \textbf{0.314} & 11.6 \\
 \hline
 \\[-1em]
 $\checkmark$ & $\checkmark$ & \textbf{0.768} & \textbf{1.408} & \textbf{0.973} & \textbf{0.032} & 0.309 & \textcolor{red}{\textbf{15.2}} \\
 \hline
\end{tabular}
\caption{Ablation studies. Best result is shown with bold. A noticeable decrease in quality is shown in red.}
\label{table:2}
\end{table}

{\small
\bibliographystyle{ieee_fullname}
\bibliography{egbib}
}

\end{document}